# On the Philosophical, Cognitive and Mathematical Foundations of Symbiotic Autonomous Systems (SAS)

Yingxu Wang[1*], Fakhri Karray[2], Sam Kwong[3], Konstantinos N. Plataniotis[4], Henry Leung[5], Ming Hou[6], Edward Tunstel[7], Imre J. Rudas[8], Ljiljana Trajkovic[9], Okyay Kaynak[10], Janusz Kacprzyk[11], Mengchu Zhou[12], Michael H. Smith[13], Philip Chen[14], and Shushma Patel[15]

[1, 5] FIEEEs, Dept. of Electrical & Computer Engineering
Schulich School of Engineering and Hotchkiss Brain Institute
Int'l Institute of Cognitive Informatics & Cognitive Computing (ICICC)
University of Calgary, Canada
Emails: yingxu@ucalgary.ca and leungh@ucalgary.ca

[2] FIEEE, Dept. of Electrical & Computer Engineering
University of Waterloo, ON, Canada
Email: karray@uwaterloo.ca

[3] FIEEE, Department of Computer Science
City University of Hong Kong, Hong Kong
Email: cssamk@cityu.edu.hk

[4] FIEEE, Dept. of Electrical & Computer Engineering
University of Toronto, ON, Canada
Email: kostas@ece.utoronto.ca

[6] SMIEEE, Toronto Research Centre, DRDC, Canada
Email: ming.hou@drdc-rddc.gc.ca

[7] FIEEE, Autonomous & Intelligent Systems Dept.
*Raytheon Technologies Research Center, USA*
Email: tunstel@ieee.org

[8] FIEEE, University Research and Innovation Center (EKIK)
Óbuda University, Budapest, Hungary
Email: rudas@uni-obuda.hu

[9] FIEEE, School of Engineering Science, Simon Fraser University
Burnaby BC, Canada
Email: ljilja@cs.sfu.ca

[10] FIEEE, Bogazici University, Bebek, Istanbul, Turkey
Email: okyay.kaynak@boun.edu.tr

[11] FIEEE, Systems Research Institute, Polish Academy of Sciences
Warsaw, Poland
Email: kacprzyk@ibspan.waw.pl

[12] FIEEE, Dept. of Computer Science, New Jersey Institute of Technology
NJ., USA
Email: mengchu.zhou@njit.edu

[13] SMIEEE, Furaxa, Inc., Orinda, CA, USA
Email: m.h.smith@ieee.org

[14] FIEEE, School of Computer Science and Engineering
South China University of Technology, Guangzhou, China
Email: philip.chen@ieee.org

[15] FBCS, Faculty of Computing, Engineering and Media
De Montfort University, Leicester LE1 9BH, UK
Email: shushma@dmu.ac.uk

## Abstract

Symbiotic Autonomous Systems (SAS) are advanced intelligent and cognitive systems exhibiting autonomous collective intelligence enabled by coherent symbiosis of human-machine interactions in hybrid societies. Basic research in the emerging field of SAS has triggered advanced general AI technologies functioning without human intervention or hybrid symbiotic systems synergizing humans and intelligent machines into coherent cognitive systems. This work presents a theoretical framework of SAS underpinned by the latest advances in intelligence, cognition, computer, and system sciences. SAS are characterized by the composition of autonomous and symbiotic systems that adopt bio-brain-social-inspired and heterogeneously synergized structures and autonomous behaviors. This paper explores their cognitive and mathematical foundations. The challenge to seamless human-machine interactions in a hybrid environment is addressed. SAS-based collective intelligence is explored in order to augment human capability by autonomous machine intelligence towards the next generation of general AI, autonomous computers, and trustworthy mission-critical intelligent systems. Emerging paradigms and engineering applications of SAS are elaborated via an autonomous knowledge learning system that symbiotically works between humans and cognitive robots.

## 1. Introduction

Philosophy is a formal means of thought for abstraction and inference that not only enables inductive theories of knowledge to be developed based on real-world observations, but also constitutes deductive inferences for rigorous thinking [1, 2, 19]. Human intelligence and wisdom are overarchingly represented by the cognitive and scientific philosophy. Basic research and engineering demands on autonomous and symbiotic systems embody a contemporary philosophy of the extended understanding towards intelligence and knowledge sciences in the era from information to intelligence revolution. Symbiotic autonomous systems are an emerging field of general AI methodology underpinned by the latest advances in intelligence, cognition, computer, and system sciences. The term *symbiosis* indicates mutually coherent and heterogeneously intelligent systems for enabling collective intelligence by autonomous human-machine interactions in the emerging hybrid society where humans and smart machines work symbiotically [3-6, 39, 46].

A symbiotic worldview, as illustrated in Fig. 1, is an overarching abstraction of the universe of discourse of nature. In Fig. 1, the double arrows denote bi-directional relations between the essences in the dual world where known relations are denoted by solid lines and relations yet to be revealed are denoted by dashed lines.

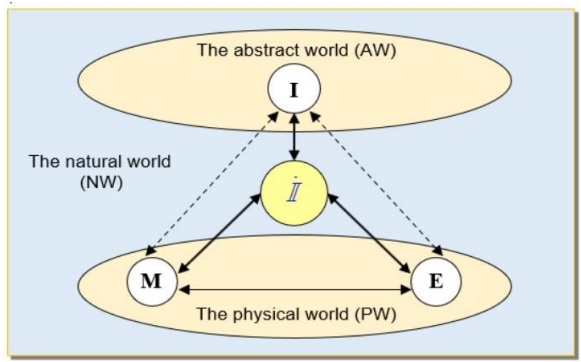

Fig. 1. The universe of discourse of the symbiotic worldview

**Definition 1.** The *universe of discourse of a symbiotic worldview* is a dual denoted by the information-matter-energy-intelligence (IME-I) model of the natural world (NW). One facet of NW is the physical world (PW) that is modeled by *matter* (M) and *energy* (E) while the other facet is the abstract world (AW) that is represented by *information* (I) as a generic representation of human perceptions and abstractions. In the IME-I model, *intelligence* ($𝕀$) plays a central role for the transformation among I, M, and E.

The latest development in SAS is underpinned by the deep explorations in intelligence science and the fast growing demand for symbiotic societies where collective intelligence of humans and machines may coherently work together. It is recognized that the universe of discourse of sciences may be categorized into natural and abstract sciences [7] as shown in Fig. 2 which indicates the extended horizon of human knowledge, the deepened understanding of the natural world, and their symbiotic interactions.

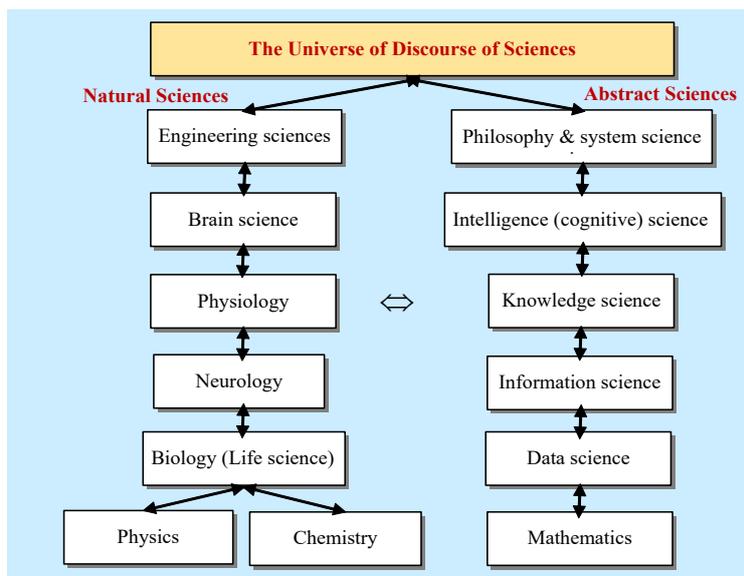

Fig. 2. The universe of discourse of contemporary sciences

The universe of discourse of contemporary sciences indicates that human abstraction and reasoning power have been advancing from concrete to abstract sciences. The concrete sciences are scientific disciplines about natural and material entities including physics, chemistry, biology, neurology, physiology, brain, and engineering sciences. However, the abstract sciences are scientific disciplines about the abstract entities including data, information, knowledge, intelligence, mathematics, philosophy, and system sciences. This leads to the findings in symbiotic cognitive cybernetics where intelligence may not be directly aggregated from data no matter how big they are, because there are multiple inductive layers from data to intelligence [7].

**Definition 2.** *Intelligence* $\mathbb{I}$ is a human, animal, or system ability that autonomously transfers a piece of information $I$ into a behavior $B$ or an item of knowledge $K$:

$$\mathbb{I} \triangleq f_{to-do} : I \to B \quad (a) \\ \parallel f_{to-be} : I \to K \quad (b) \tag{1}$$

where Eq. (1.a) denotes the narrow sense of intelligence, while Eq. (1.a) and Eq. (1.b) in parallel ($\parallel$) represent the broad sense of intelligence.

A series of white papers on SAS have been released by the IEEE Symbiotic Autonomous Systems Initiative (SASI) [48], which indicates the importance of the topic recognized by IEEE, a worldwide opinion making society. IEEE SASI has recognized a wide range of SAS applications including ambient technologies, intelligent interactions, adaption to environment, robotized objects, smart cities, industrial robots, augmented humans, ethical challenges, and supply-driven economy. One of the findings by SASI is the crucial need for developing a formal approach to SAS rooted in and based on various fields of science and technology exemplified by cognitive science, computer science, robotics, autonomous decision-making system [5, 15, 19, 37, 38, 39, 46] and run-time intelligent behavioral generation [40]. It is recognized that such a formally sound theory would accelerate the development of the field and facilitate its practical applications.

The advances towards SAS are driven by the theoretical and technological development across intelligence science, brain science, cognitive science, robotics, and computational intelligence. Basic studies of the fundamental constraints and challenges of intelligence science and AI have led to the breakthroughs towards SAS beyond adaptive systems by conventional deterministic computing technologies. SAS is expected to address the persistent challenges of nondeterministic decision-making, autonomous intelligence generation, and run-time machine behavioral generation beyond stored-program-based computing [8]. SAS that involve human collaborations (or interventions) to complete some highly intelligent tasks, become the essence in future technologies and applications [37].

This work explores the philosophical, cognitive and mathematical foundations of SAS. It investigates the technical bottlenecks of SAS and indispensable theories underpinned by the latest advances in intelligence science, computational intelligence, and intelligent mathematics. It addresses the challenge to seamlessly incorporate human and machine interactions in SAS theoretically and pragmatically. This may enhance SAS-based machine intelligence towards autonomous and cognitive intelligence in order to augment human collective capability by advanced machine intelligence. Sections 2 and 3 formally elaborate symbiotic and autonomous systems, respectively, which lead to the theoretical framework of SAS in Section 4. Emerging SAS paradigms including machine knowledge learning systems, brain-inspired systems, and cognitive robots are demonstrated. The breakthroughs in SAS are expected to trigger a wide range of novel applications towards the next generation of general AI, autonomous computers, and mission-critical intelligent systems.

## 2. Symbiotic Systems

Symbiosis is a widely observable phenomenon in biological, mental, and social systems where mutual dependences exist among plants, animals, and human societies as a necessary condition for them to co-evolve [3-6]. Symbiosis is particularly important to human societies because of the fundamental need for extending individuals' physical, intellectual, and/or resource limits. Therefore, it becomes a fundamental principle of system science and the universal context of modern sciences and engineering.

**2.1 Conceptual Model of Symbiotic Systems**

**Definition 3.** A *symbiotic system* (*SS*) is a bio-brain-social-inspired system characterized by heterogeneously coherent autonomous structures and behaviors for embodying collective intelligence.

Recent basic research and theoretical breakthroughs [5] have led to the systematic revelations of the parallel and recursive framework of contemporary sciences and human-nature symbiosis. In a broad sense, all subsystems of a complex system are symbiotic in the forms of positive and negative coherency where the former is for structural or functional augmentation of the system, while the latter is for system stabilization and autonomous control [40].

An example of a natural *SS* with negative feedback is the Predator-Prey System (PPS) where the populations of lions and gazelles in a territory are modeled as a dynamic *SS* by a system of ordinary differential equations:

$$PPS \triangleq \begin{cases} \dfrac{dN_L}{dt} = b_L N_L N_G - d_L N_L \\ \dfrac{dN_G}{dt} = b_G N_G - d_G N_G N_L \end{cases} \quad (2)$$

where $N_L$ and $N_G$ represent the populations of lions and gazellas, $b_L$, $b_G$, $d_L$, and $d_G$, the birth and death rates of them, respectively.

The simulation of this natural *SS* in MATLAB is shown in Fig. 3 where the predator and prey are mutually dependent for food and healthy, respectively. It is interesting in this natural symbiotic eco-system, the population of the predators are eventually controlled by that of the preys.

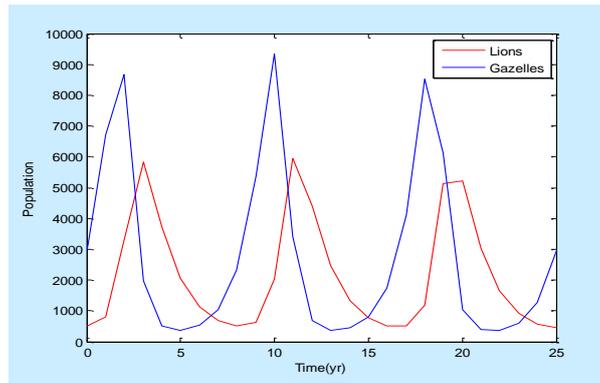

Fig. 3. Simulation of a predator-prey symbiotic system

**Definition 4.** The mathematical model of a *general SS* § is an 8-tuple in the universe of discourse $\mathfrak{U}$:

$$\S \triangleq (C, B, R^c, R^b, R^f, R^i, R^o, \Theta) \quad (3)$$

where $C$ is a finite set of *components* of §; $B$ a finite set of *behaviors*; $R^c = C \times C$ a set of component relations; $R^b = B \times B$ a set of behavioral relations; $R^f = B \times C$ a set of functional relations; $\Theta$ the environment of § as equivalent but external *SS*; $R^i = \Theta.B \times \S.B$ a set of input relations between the *B* dimension of $\Theta$ and *S* (i.e., $\Theta.B$ and $\S.B$); and $R^o = \S.B \times \Theta.B$ a set of output relations.

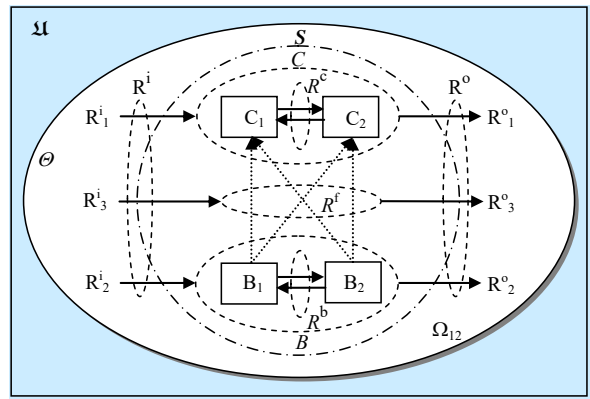

Fig. 4. The hyperstructural model of a general symbiotic system

Definition 4 provides a formal model of general *SS* that constitutes the generic properties of arbitrary *SS*. The structure of the abstract *SS* defined in Definition 4 is illustrated in Fig. 4 where *C*, *B*, and $\mathcal{R} = \{R^c, R^b, R^f, R^i, R^o, \Theta\}$ denote the components, behaviors, and their structural/behavioral/functional/input/output relations, respectively. The environment $\Theta$ represents the external space of the given *SS* by other related *SS'*.

According to Definition 4, human group coordination theory, sociology, and management science are an applied *SS* domain. It is discovered that *coherency* is the essence of *SS* [17-19] in *SS* theory, social networks, and IoTs. For instance, in computer science and complex software engineering, a mechanism is found that the number of people in a coordinated group may be traded with time or project duration under certain symbiotic conditions [20].

**2.2 Topological Properties of Symbiotic Systems**

*SS* is pervasively indispensable in the natural world in general and human societies in particular. The general topology of an *SS* or a system of *SS'* is a hierarchical and recursive structure rather than a flat one [18, 19] based on the conceptual model of *SS*.

**Definition 5.** The *general topology* $\hat{S}$ of *SS* is a recursively embedded structure where each *k*th layer $S^k$ in the *n*-layer system hierarchy is embedded into the adjacent upper layer $S^{k+1}$:

$$\hat{S} \triangleq \underset{k=1}{\overset{n}{R}} S^k(S^{k-1}), \quad S^0 = (C_0, B_0, R_0^c, R_0^b, R_0^f, R_0^i, R_0^o, \Theta_0) \tag{4}$$
$$= S^n(S^{n-1}(...(S^1(S^0))))$$

where the *big-R notation* [20] is a general recursive operator that denotes a set of recurring structures or a set of iterative/embedded behaviors.

**Theorem 1.** The *general topology* $\hat{S}$ of *SS* is constituted by the following properties:
  a) $\hat{S}$ is *necessarily* constrained by bottom-up inductions; and
  b) $\hat{S}$ is *sufficiently* constrained by top-down deductions.

**Proof.** $\forall S^0 = (C^0, B^0, \mathcal{R}^0, \Theta_0) = (C_0, B_0, R_0^c, R_0^b, R_0^f, R_0^i, R_0^o, \Theta_0)$ as a primitive *SS* according to Definition 5 where all attributes are known and concrete:

a) The first property of $\hat{S}$ holds by a series of bottom-up inductions *iff* $S^0$ is given:

$$\begin{aligned}
S^1 &= \underset{i_1=1}{\overset{k_1}{R}} S_{i_1}^0 = \underset{i_1=1}{\overset{k_1}{R}} (C_{i_1}^0, B_{i_1}^0, \mathcal{R}_{i_1}^0, \Theta_{i_1}^0) \\
S^2 &= \underset{i_2=1}{\overset{k_2}{R}} S_{i_2}^1 = \underset{i_2=1}{\overset{k_2}{R}} \underset{i_1=1}{\overset{k_1}{R}} S_{i_2 i_1}^0 = \underset{i_2=1}{\overset{k_2}{R}} \underset{i_1=1}{\overset{k_1}{R}} (C_{i_2 i_1}^0, B_{i_2 i_1}^0, \mathcal{R}_{i_2 i_1}^0, \Theta_{i_2 i_1}^0) \\
&\cdots \\
S^n &= \underset{i_n=1}{\overset{k_n}{R}} \cdots \underset{i_2=1}{\overset{k_2}{R}} \underset{i_1=1}{\overset{k_1}{R}} S_{i_n...i_2 i_1}^0 = \underset{i_n=1}{\overset{k_n}{R}} \cdots \underset{i_2=1}{\overset{k_2}{R}} \underset{i_1=1}{\overset{k_1}{R}} (C_{i_n...i_2 i_1}^0, B_{i_n...i_2 i_1}^0, \mathcal{R}_{i_n...i_2 i_1}^0, \Theta_{i_n...i_2 i_1}^0) \\
&= S^n(S^{n-1}(...(S^1(S^0)))) = \hat{S}
\end{aligned} \tag{5}$$

b) The second property of $\hat{S}$ holds by a series of top-down deductions as an inversed process of Eq. (5) that reduce the *SS* from top-down until it is realized by $S^0$.

∎

According to Theorem 1, a set of properties of system symbiosis may be derived including: a) The *general topological structure of SS* is a recursive hierarchical structure that links the embedded structures and relations between two arbitrarily adjacent layers of *SS*; b) The *abstraction principle of systems* states that any *SS* may be inductively integrated and composed with decreasing details at different layers, $0 \leq k \leq n$, from the bottom up; and c) The *refinement principle of systems* states that any *SS* may be deductively specified and analyzed with increasing details at different layers, $0 \leq k \leq n$, from the top down.

**2.3 Symbiotic Mechanisms of Symbiotic Systems**

The principle of symbiosis among systems may be formally described by the mechanism of system fusions formally expressed by a novel mathematical model of *incremental union* of sets of relations [18]. In this approach, the symbiotic system gains from simple standalone or lower-level systems may be rigorously explained and quantitatively determined.

**Definition 6.** A *symbiotic fusion* ⊞ is an incremental union of a pair of sets of relations, $R_1$ and $R_2$, between two systems $S_1$ and $S_2$:

$$R(S_1, S_2) \triangleq R_1 \boxplus R_2 = R_1 \cup R_2 \cup \Delta R_{12}(C_1, C_2) \tag{6}$$

where $C_1$ and $C_2$ are sets of components (entities) in $S_1$ and $S_2$, respectively.

Based on Definition 6, an important *SS* principle known as the symbiotic mechanism is introduced to explain the nature of system symbiosis.

**Theorem 2.** Given a pair of arbitrary systems $S_1 = (C_1, B_1, \mathcal{R}_1, \Theta_1)$ and $S_2 = (C_2, B_2, \mathcal{R}_2, \Theta_2)$, *system symbiosis* $\Omega_s(S_1, S_2)$ in an *SS* is generated by a set of symbiotic gains $\Delta R_{12}$ during a symbiotic fusion ⊞:

$$\Omega_s(S_1, S_2) = |\Delta R_{12}(C_1, C_2)| \\ = 2|C_1| \cdot |C_2|, \ \forall R \triangleq R_1 \boxplus R_2 = R_1 \cup R_2 \cup \Delta R_{12} \tag{7}$$

where the newly created symbiotic gain $\Delta R_{12} \subset R_1 \boxplus R_2$ but $\Delta R_{12} \not\subset R_1 \wedge \Delta R_{12} \not\subset R_2$.

**Proof.** $\forall S_1 = (C_1, B_1, \mathcal{R}_1, \Theta_1)$ and $S_2 = (C_2, B_2, \mathcal{R}_2, \Theta_2)$, the symbiotic result $\Delta R_{12}$ are obtained as the difference between the size of the entire relations $|R|$ of the *SS* and the sum of those of the individual subsystems $|R_1|$ and $|R_2|$. Therefore, Theorem 2 holds because:

$$\forall \Delta R_{12} \sqsubset SS, \ R_1 \sqsubset S_1, \text{ and } R_2 \sqsubset S_2, \\ \Omega_s(S_1, S_2) = |\Delta R_{12}(C_1, C_2)| \\ = |R| - (|R_1| + |R_2|) \\ = (n_{c_1} + n_{c_2})^2 - (n_{c_1}^2 + n_{c_2}^2) = (n_{c_1}^2 + 2n_{c_1}n_{c_2} + n_{c_2}^2) - (n_{c_1}^2 + n_{c_2}^2) \\ = 2n_{c_1}n_{c_2} \\ = 2|C_1||C_2| \tag{8}$$

∎

Theorem 2 formally elaborates the natural mechanisms of symbiosis in *SS* and how it is determined rigorously. The symbiotic principle of *SS* as derived in Theorem 2 explains that the composition ⊎ of two systems, $SS = S_1 \uplus S_2$, results in the generation of new functions $\Delta R_{12}$ that solely belong to the *SS* but do not exist in any of the individual systems $S_1$ or $S_2$ when they stand alone.

For instance, given a pair of simple systems $S_1$ and $S_2$ as shown in Fig. 5, the symbiotic system generated by their composition results in a new set of relations or functions, as well as complexities, that may be rigorously determined according to Theorem 2 as $\Omega_{12} \cdot |\Delta R_{12}| = 2|C_1| \cdot |C_2| = 2(3 \cdot 2) = 12$.

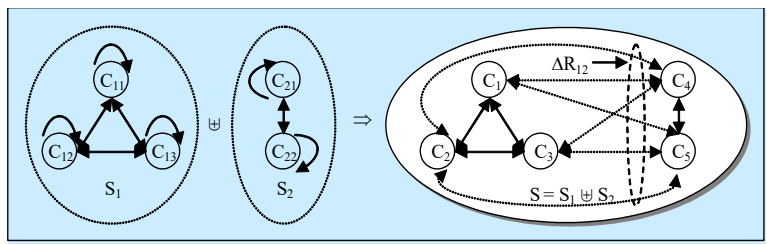

Fig. 5. The symbiosis of general systems and the SAS gains

Another symbiotic principle of group gains may be formally described based on the *random nature of human-system errors* in group task performing [45]. A statistical property of *SS* is that the occurrences of an identical error by different individuals may most likely happen at different times. This human error-making mechanism in symbiotic groups provides a foundation for fault-tolerance towards *symbiotic trustworthiness*. It indicates that human/system errors may be prevented from happening or be collectively cancelled in a symbiotic architecture of peers and/or human-machine systems. The next result states the principle of symbiotic error cancellation.

**Theorem 3.** *Symbiotic error cancellation* in an *SS* is a mechanism that the collective reliability of an *SS*, $\mathfrak{N}(SS)$, as a complement of its error rate $\mathfrak{E}(SS)$, is always greater than the sum of its *n* individual subsystems $\mathfrak{N}(\sum_{k=1}^{n} S(k))$:

$$\mathfrak{N}(SS) \gg \mathfrak{N}(\sum_{k=1}^{n} S(k)) \tag{9}$$

where $\mathfrak{N}(SS)$ is the symbiotic trustworthiness gained by the *SS*.

***Proof.*** Let the symbiotic error cancellation $\triangle_r(SS) = \mathfrak{N}(SS) - \mathfrak{N}(\sum_{k=1}^{n} S(k))$ be the difference between the reliabilities of the *SS* and the sum of its subsystems *S(k)*.

$$\begin{aligned}
\forall \, \mathfrak{N} &= 1 - \mathfrak{E} \text{ and } \underset{k=1}{\overset{n}{R}} (r_e(k) < 1.0), \\
\triangle_r(SS) &= \mathfrak{N}(SS) - \mathfrak{N}(\sum_{k=1}^{n} S(k)) \\
&= (1 - \prod_{k=1}^{n} r_e(k)) - (1 - \sum_{k=1}^{n} r_e(k)) \\
&= \sum_{k=1}^{n} r_e(k) - \prod_{k=1}^{n} r_e(k), \; \underset{k=1}{\overset{n}{R}} (r_e(k) < 1.0) \\
&\gg 0 \\
\Rightarrow \mathfrak{N}(SS) &\gg \mathfrak{N}(\sum_{k=1}^{n} S(k))
\end{aligned} \tag{10}$$

∎

Empirically, the symbiotic reliability of a system may be gained by intensive rechecking based on the random nature of error distributions and the independent nature of error patterns among individuals or components in an *SS*. The principle of symbiotic error cancellation may be applied in *SS* and hybrid intelligent systems where humans and machines work together, particularly in mission-critical and safety-critical *SS*.

It is recognized that humans are the most dynamic and active part of *SS*. Because the most matured *SS* paradigm is the brain, advanced *SS* are naturally open to incorporate human intelligence as indicated in Fig. 6. According to Theorems 2 and 3, a hybrid *SS* with humans in the loop gains strengths towards the implementation of cognitive intelligent systems. The cognitive *SS* sufficiently enables a powerful intelligent system with the strengths of both human and machine intelligence. This is why intelligence and system sciences may inspire *SS* designers to develop fully autonomous intelligent systems in highly demanded engineering applications. More general *SS* gains have been recognized in this work including some that: a) Extend physical capability of individuals; b) Improve group productivity and efficiency; c) Enhance quality and reliability; d) Enable information/skills/knowledge share; e) Gain exponential learning power; and f) Enable collective intelligence and wisdom.

*SS* Paradigms include simulated natural intelligence systems, social computing systems, human-machine systems, cognitive systems, cognitive robots, bioinformatics systems, brain-inspired systems, self-driving automobiles, unmanned systems, and intelligent IoT. Some of them are analyzed in Section 4.

## 3. Autonomous Systems

The transdisciplinary advances in intelligence, cognition, computer, cybernetic, and systems sciences have led to the emerging field of autonomous systems [9-14, 33, 41, 42]. The ultimate goal of autonomous systems is to implement a brain-inspired system that may think and act as a human counterpart in hybrid intelligent systems. Various autonomous systems are demanded to address the pertinacious challenges to classic AI due to the lack of cognitive, intelligent, computational, and mathematical readiness for real-time and training-free applications [8, 16].

**Definition 7.** *Autonomous systems* (*AS*) are advanced intelligent systems that function without human intervention for implementing complex cognitive abilities aggregating from reflexive, imperative, and adaptive intelligence to autonomous and cognitive intelligence.

## 3.1 Intelligence Science Foundations of Autonomous Systems

Intelligence is the paramount cognitive ability of humans that may be mimicked by *AS* implemented by computational intelligence. A classification of intelligent systems may be derived according to the forms of system inputs and outputs as shown in Table 1. The low-level reflexive and imperative systems are able to process deterministic stimuli by deterministic or indeterministic algorithms. The adaptive systems are designed to deal with indeterministic stimuli by deterministic behaviors predefined at the design time. However, *AS* is characterized by both indeterministic stimuli and indeterministic behaviors pending for run-time contexts.

Table 1. Classification of autonomous and nonautonomous systems

| **Intelligent behaviors** | | **Behavior (O)** | |
|---|---|---|---|
| | | Deterministic | Indeterministic |
| **Stimulus (I)** | Deterministic | *Reflexive* | *Imperative* |
| | Indeterministic | *Adaptive* | *Autonomous* |

**Definition 8.** *Intelligence science* is a contemporary discipline that studies the mechanisms and properties of intelligence, formal principles, and mathematical means of intelligence, the generation of intelligence through the neural, cognitive, functional, and mathematical levels, as well as their engineering and computational implementations.

A *Hierarchical Intelligence Model* (HIM) is introduced to reveal the levels of intelligence and their increasing complexities and difficulties for implementation in computational intelligence in Fig. 6 based on the *abstract intelligence* (αI) theory [8]. In HIM, the levels of intelligence are aggregated from reflexive, imperative, adaptive, autonomous, and cognitive intelligence. Types of system intelligence across the HIM layers may be formally described by using the pattern of stimulus/event-driven behaviors that will be formally modeled in Definition 8.

The field of *AS* studies the properties of intelligence for realizing brain-inspired systems by high-level machine intelligence beyond those of imperative and adaptive systems with deterministic or prescriptive behaviors. In the past 60 years of AI and systems engineering, few actual *AS*' were developed because the theoretical foundations for autonomous intelligence and systems have not matured sufficiently [8, 16]. Many current AI systems are still bounded by the intelligence bottleneck of adaptive mechanisms where machine intelligence is constrained by the lower-level reflexive, imperative, and deterministic intelligent abilities. The findings indicate that, to extend the intelligence power of traditional AI systems via *AS*, a *general AI* (GAI) system needs not only to mimic the imperative and iterative intelligence, but also to realize more powerful human equivalent intelligence according to the HIM theory.

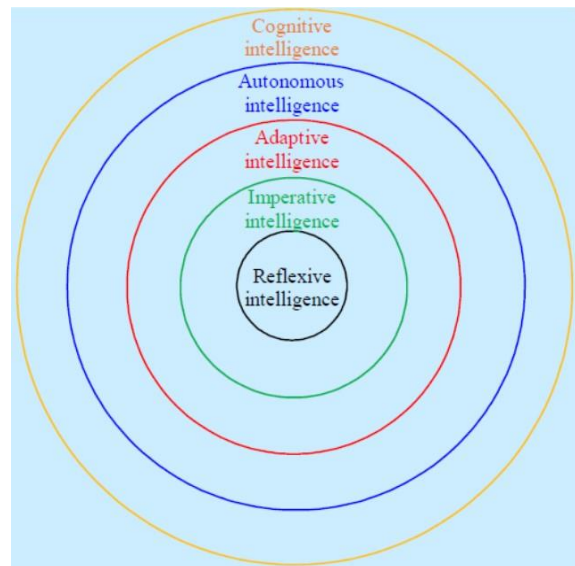

Fig. 6. The hierarchical intelligence model (HIM) of AS

## 3.2 System Science Foundations of Autonomous Systems

The structural properties of *AS* naturally fit the principle of recursive systems [17-19, 46] very well. According to the HIM model, the *AS* framework provides an explanation of how the advanced AS intelligence is generated from the bottom up based on the transformability among the five-level hierarchy of system intelligence. The HIM framework of *AS* outlined in Fig. 6 may be refined by 16 intelligent behaviors at five levels listed in Table 2. Each of the intelligent behaviors may be formally described according to the *event dispatching* operation in [8].

**Definition 9.** The *mathematical model of AS* is a formalization of HIM by a unified event-driven dispatching pattern that denotes all 16 intelligent behaviors at five levels:

$$AS \triangleq \underset{i=1}{\overset{5}{R}} \underset{j=1}{\overset{m_i}{R}} [@e_j^i | \mathbb{T} \hookrightarrow B_j^i | \text{PM}] \tag{11}$$

where $\underset{i=1}{\overset{5}{R}} m_i = [1,3,3,5,4]$, $\hookrightarrow$ represents the *event dispatching* operator, $@e_j^i | \mathbb{T}$ the *j*th event prespecified by a type suffix $|\mathbb{T}$ at the *i*th level, and $B_j^i | \text{PM}$ the corresponding behavior triggered by a certain event as a process model (PM).

Table 2. The Framework of Autonomous Intelligence Embodied in AS

| Level | Category | Type | Symbol | Description |
|---|---|---|---|---|
| 1 | Reflective intelligence | Reflexive | $\mathbb{I}_{ref}$ | A wired behavior directly driven by specifically coupled external stimuli or triggering event |
| 2 | Imperative intelligence | Event-driven | $\mathbb{I}_{imp}^e$ | A predefined imperative behavior driven by an event |
| | | Time-driven | $\mathbb{I}_{imp}^t$ | A predefined imperative behavior driven by a point of time |
| | | Interrupt-driven | $\mathbb{I}_{imp}^{int}$ | A predefined imperative behavior driven by a system- triggered interrupt event |
| 3 | Adaptive intelligence | Analogy-based | $\mathbb{I}_{adp}^{ab}$ | An adaptive behavior that operates by seeking an equivalent solution for a given analog request |
| | | Feedback-modulated | $\mathbb{I}_{adp}^{fm}$ | An adaptive behavior rectified by the feedback of temporal system output |
| | | Environment-aware | $\mathbb{I}_{adp}^{ea}$ | An adaptive behavior where multiple prototype behaviors are modulated by the change of external environment |
| 4 | Autonomous intelligence | Perceptive | $\mathbb{I}_{aut}^{pe}$ | An autonomous behavior based on the selection of a perceptive inference |
| | | Problem-driven | $\mathbb{I}_{aut}^{pd}$ | An autonomous behavior that seeks a rational solution for a given problem |
| | | Inference-driven | $\mathbb{I}_{aut}^{id}$ | An autonomous behavior seeking an optimal path towards the given goal |
| | | Decision-driven | $\mathbb{I}_{aut}^{dd}$ | An autonomous behavior embodied by the outcome of a decision process |
| | | Deductive | $\mathbb{I}_{aut}^{de}$ | An autonomous behavior driven by a deductive process based on known principles |
| 5 | Cognitive intelligence | Knowledge-based | $\mathbb{I}_{cog}^{kb}$ | A cognitive behavior generated by introspection of acquired knowledge |
| | | Learning-driven | $\mathbb{I}_{cog}^{ld}$ | A cognitive behavior generated by both internal introspection and external acquisition |
| | | Goal-driven | $\mathbb{I}_{cog}^{gd}$ | A cognitive behavior that creates a causal chain from a problem to a rational solution |
| | | Inductive | $\mathbb{I}_{cog}^{id}$ | A cognitive behavior that draws a general rule based on multiple observations or common properties |

*AS* exhibits nondeterministic, context-dependent, and adaptive behaviors of machine intelligence. *AS* is a nonlinear system that not only depends on current stimuli or demands to the system, but is also modulated by internal status and perceptions formed by historical events and current rational goals.

**Theorem 4.** An *AS* is characterized by a) a *hierarchical* architecture and b) a series of *recursively inclusive* behaviors, i.e.:

$$AS|_\S \triangleq \begin{cases} a) \underset{k=1}{\overset{4}{R}} B^k(B^{k-1}), B^0 = \underset{i=1}{\overset{n_{ref}}{R}} @e_i | \text{REF} \hookrightarrow B_{ref}(i) | \text{PM} \\ b) B_{Cog} \supseteq B_{Aut} \supseteq B_{Ada} \supseteq B_{Imp} \supseteq B_{Ref} \end{cases} \tag{12}$$

**Proof.** a) The behavioral architecture $\underset{k=1}{\overset{4}{R}} B^k(B^{k-1})$ of an arbitrary *AS* in Eq. (12.a) aggregates $B^0$ through $B^4$ hierarchically from the bottom up *iff* $B^0$ is deterministic. b) Because the five levels of AS behaviors in Eq. (12.b) is a partial order across all layers, they are *recursively inclusive* from the bottom up. ∎

Theorem 4 indicates that any lower layer *AS* behavior is a subset of those of a higher layer. In other words, any higher layer *AS* behavior is a natural aggregation of those of lower layers as shown in Eq. (12) and Fig. 6.

The theories and technologies of *AS* explain the evolution of human and system intelligence as an inductive process of intelligence generation. They also reveal the properties of system autonomy at five levels implemented by computational intelligence and system engineering technologies. Advances in *AS* are expected to pave a way towards highly intelligent machines for augmenting human capabilities. Typical emerging *AS* include unsupervised computational intelligence, cognitive systems, brain-inspired systems, general AI, cognitive robots, unmanned systems, human intelligence augmentation systems, and intelligent IoTs.

## 4. Symbiotic Autonomous Systems

The synergy between *AS* and *SS* has logically led to the emergence of symbiotic autonomous systems [3-6]. As advanced intelligent system theories and technologies, they enable *AS*, *SS*, and human groups to work coherently in a hybrid society towards the generation of collective intelligence. It leads to a unified universe of discourse of cybernetics underpinned by the emergence of abstract sciences and their counterparts of classical concrete sciences as introduced in Fig. 2.

It is recognized that epistemology plays an indispensable role in the kernel of philosophy for human intelligence expression and advancement [2, 21]. A key approach to system intelligence augmentation by symbiotic autonomous systems is via symbiotic knowledge learning for autonomous problem solving. Knowledge learning helps to extend the state space of mind as the foundation for creativity and wisdom generation; while autonomous problem solving enables a symbiotic autonomous system to generate intelligent behaviors at run-time as a counterpart of humans in the loop [8, 10, 43, 44].

**Definition 10.** *Symbiotic Autonomous Systems* (*SAS*) are advanced intelligent and cognitive systems embodied by computational intelligence in order to facilitate collective intelligence among human-machine interactions in a hybrid society.

### 4.1 Conceptual Model of SAS

Recent basic research of *SAS* has revealed that novel solutions to fundamental AI problems are deeply rooted in the understanding of both natural intelligence and maturity of suitable mathematical tools for rigorously modeling the brain in machine understandable forms [7]. It leads to the emergence of Intelligent Mathematics (*IM*) [22, 23] as a category of the denotational mathematics as a supplement to the inadequate power of classic analytic and numerical mathematics. *IM* is demanded by *SAS* based on the observation that the domain of the cognitive entities has been out of the traditional domain of real numbers ($\mathbb{R}$). As a result, new problems demand new mathematics recognized as *IM* for dealing with those of hyperstructures ($\mathbb{H}$) beyond that of classic mathematics of logic, numerical methods, probability, and analytics.

**Definition 11.** The *hierarchy of cognitive objects CO* represented in the brain as typical $\mathbb{H}$ is a 4-tuple in the categories of *data* ($\mathbb{D}$), *information* ($\mathbb{I}$), *knowledge* ($\mathbb{K}$), and *intelligence* ($\dot{\mathbb{I}}$) from the bottom up according to their levels of abstraction:

$$\mathcal{CO} \triangleq (\mathbb{D}, \mathbb{I}, \mathbb{K}, \dot{\mathbb{I}}) = \begin{cases} \mathbb{D} = f_d : O \to Q \\ \mathbb{I} = f_i : D \to S \\ \mathbb{K} = f_k : I \to C \\ \dot{\mathbb{I}} = f_{\dot{i}} : I \to B \end{cases} \quad (13)$$

where the symbols denote the cognitive entities *object* (*O*), *quantity* (*Q*), *semantics* (*S*), *concept* (*C*), and *behavior* (*B*), respectively.

The relationship among the cognitive entities in the brain may be formally described by a hierarchical model according to the SS theory of Theorem 1. It derives the following principle for explaining how intelligence is generated in the brain from low-level entities.

**Corollary 1.** Let $\kappa^0$ through $\kappa^4$ be the hierarchical layers of human cognition objects of data ($\mathbb{D}$), information ($\mathbb{I}$), knowledge ($\mathbb{K}$), and intelligence ($\dot{\mathbb{I}}$). The *transformability* among the cognitive entities is determined by a recursive structure $S_\Xi$ where any higher-layer system or entity is represented by that of its lower layers deductively, or vice versa inductively:

$$S_\Xi = \mathop{R}_{k=1}^{4} \kappa^k(\kappa^{k-1}), \quad \kappa^0 = \mathop{R}_{i=0}^{n} d_i|\mathbb{T}_i$$
$$= \kappa^4(\kappa^3(\kappa^2(\kappa^1(\kappa^0))))$$
$$= \dot{I}(\mathbb{K}(I(\mathbb{D}(\mathop{R}_{i=0}^{n} d_i|\mathbb{T}_i)))) \tag{14}$$

Corollary 1 provides a general system science theory. Each cognitive layer ($\kappa^k$) in Eq. (14) is specified by the adjacent lower layer ($\kappa^{k-1}$) until the cognitive hierarchy is terminated at the bottom sensorial layer ($\kappa^0$), which is embodied by a set of $n+1$ dimensional data, $\mathop{R}_{i=0}^{n} d_i|\mathbb{T}_i = (d_0|\mathbb{T}_0, d_1|\mathbb{T}_1, ..., d_n|\mathbb{T}_n)$ as acquired by abstraction and quantification of the brain where a type suffix convention $|\mathbb{T}$ is adopted to denote a variable $x|\mathbb{T}$ in *SAS*.

According to Corollary 1, the philosophy of current machine learning would be questionable, because it is implemented by two phases known as *training* (domain calibration) and *reflexive regression* [24]. The former is implemented by large-scale data-driven regressions; while the latter is a reflexive classification. None of them are autonomous because the former is supervised and intensively dependent on data labeling and preprocessing. However, the latter is constrained by the pretrained domain and norms of samples as *special* solutions within a particular scope of sample data. In generic mathematical theory, there are often infinitive special solutions for the entire domain of a category of targets such as facial images and road conditions [24, 25, 35-36]. Therefore, there are no generic solutions yet discovered for all-purpose image recognition. Some systems for facial recognition may claim 80+% accuracy in USA. However, the same system may not fit those in Africa or Asia. It is surprising that any best-trained facial recognition neural network may absolutely fail when a feeding image is upside-down. These phenomena explain why so many researchers continually enter the field of facial recognition because there is a lack of general methodology in-line with the *SAS* principles and the ignorance of how the brain processes visual information [25-27].

### 4.2 Machine Knowledge Learning as a Key Paradigm of SAS

Human and machine learning as well as knowledge representation for both individuals and collective intelligence are a key *SAS* paradigm. Machine learning systems as a popular technology may have inherited fundamental immaturity and lack of rigorous theories from the *SAS* point of view. It is recognized that learning is a cognitive process of knowledge and behavior acquisition [28]. However, collective knowledge learning as the most important machine learning demand still remains a theoretical and technical challenge to GAI and *SAS*.

**Definition 12.** *Machine learning* as an *SAS* paradigm is classified into six categories encompassing: 1) Object identification; 2) Cluster classification; 3) Pattern recognition; 4) Functional regression; 5) Behavioral generation; and 6) Knowledge acquisition, as formally described as follows:

$$\begin{cases} L_i(\mathbf{x}, \mathbf{P} \mid \mathbf{x} \in \mathbf{X}) \triangleq \mathbf{x} = \mathbf{P}.\mathbf{x} & // \text{ Object identification} \\ L_c(\mathbf{X}, \mathbf{P}) \triangleq \mathbf{X} \subset \mathbf{P} & // \text{ Cluster classification} \\ L_r(\mathbf{X}, \mathbf{P}) \triangleq \mathbf{X} = \mathbf{P} & // \text{ Pattern recognition} \\ L_f(\mathbf{X}, \mathbf{P}) \triangleq \mathbf{X} \Rightarrow \mathbf{P}(\mathbf{X}) & // \text{ Functional regression} \\ L_b(\mathbf{X}, \mathbf{P}) \triangleq \mathbf{X} \Rightarrow b(\mathbf{P}(\mathbf{X})) & // \text{ Behavior generation} \\ L_k(\mathbf{X}, \mathbf{K}) \triangleq \mathbf{X} \Rightarrow c(\mathbf{X}) \uplus \mathbf{K} & // \text{ Knowledge acquisition} \end{cases} \tag{15}$$

where $X$ denotes a vector of objects $x$, $\mathbf{x} \in X$, $P$ a target set of patterns, $b$ a behavior determined by $P$, $c$ a formal concept, $K$ a set of knowledge, and $\uplus$ an operator of knowledge composition.

The last category of knowledge acquisition in Eq. (15) as revealed in [28] is underpinned by knowledge science and *IM* known as *concept* and *semantic algebras* [23]. Machine knowledge learning is indispensable because theoretically epistemology is in the center of fundamental philosophy and empirically knowledge learning is a lifelong endeavor beyond any other types of learning as identified in Definition 12.

**Definition 13.** The basic structure of *knowledge* $\kappa$ is a conceptual relation between two concepts $c_1$ and $c_2$ which is neurologically embodied by a synaptic connection between a pair of pivot neurons:

$$\kappa = c_1 \times c_2, \quad c_1, c_2 \in C \tag{16}$$

where the *formal concept* $C$ as the basic cognitive structure of knowledge is a 5-tuple:

$$C \triangleq (A, O, R^c, R^i, R^o) \tag{17}$$

which includes $A$ as a finite set of *attributes* or *intension* of $C$; $O$ a finite set of *objects* or *extension* of $C$; $R^c$ a nonempty set of *internal relations* $R^c = O \times A$; $R^i$ a set of *input relations* $R^i \subseteq C' \times C$ where $C'$ is a distinguished set of concepts in knowledge; and $R^o$ a set of *output* relations $R^o \subseteq C \times C'$.

Knowledge is one of the few fundamental abstract concepts in contemporary sciences that has yet to be accurately quantified and rigorously measurable. Empirical knowledge measurements have used to be such as "erudite," "plenty of," and "informative." However, none of them is rigorous and measurable because of the immaturity of knowledge science [28]. This is in analogy with the case that people had to use the empirical and unquantifiable unit "*horse-power*" for centuries before Newton formally clarified the unit of force is N (*newton*) [1].

Based on Definitions 13, an important discovery in basic research on the nature of knowledge by Wang [7, 29] has revealed that the basic unit of knowledge is a *binary relation* (*bir*) that is comparable to that of *bit* (binary digit) for data and information introduced by Shannon [30]. This finding is supported by empirical evidences of the synaptic connections in neurology and brain science [21].

**Definition 14.** The *basic unit of knowledge* $u(\kappa)$ is a general quantitative metrics for knowledge measurement:

$$u(\kappa) \triangleq u(c_1 \times c_2) = bir \tag{18}$$

**Theorem 5.** *Knowledge symbiosis* $\Omega_\kappa (\underset{i=1}{\overset{n}{R}} c_i, \underset{j=1}{\overset{m}{R}} c_j)$ between two sets of formal concepts results in new knowledge quantifiable by *bir*:

$$\Omega_\kappa (\underset{i=1}{\overset{n}{R}} c_i, \underset{j=1}{\overset{m}{R}} c_j) \triangleq \underset{i=1}{\overset{n}{R}}\underset{j=1}{\overset{m}{R}} (c_i \times c_j) \; [bir] \tag{19}$$

*Proof*. Theorem 5 holds based on Definition 14 by proving each of the five dimensions of attributes (intension, $A$), objects (extension, $O$), and internal/input/output relations ($R^c, R^i,$ and $R^o$):

$$\forall \underset{i=1}{\overset{n}{R}} c_i \text{ and } \underset{j=1}{\overset{m}{R}} c_j \in C = (A, O, R^c, R^i, R^o), \tag{20}$$

$$\Omega_\kappa(\underset{i=1}{\overset{n}{R}} c_i, \underset{j=1}{\overset{m}{R}} c_j) = \underset{i=1}{\overset{n}{R}}\underset{j=1}{\overset{m}{R}} (c_i \times c_j)$$

$$= \underset{i=1}{\overset{n}{R}}\underset{j=1}{\overset{m}{R}} \begin{cases} c_i.A_i \times c_j.A_j & | \; [ \; (| A_i | \bullet | A_j |) \; bir] \\ c_i.O_i \times c_j.O_j & | \; [ \; (| O_i | \bullet | O_j |) \; bir] \\ c_i.R_i^c + c_j.R_j^c & | \; [ \; (| O_1 \times A_1 | + | O_2 \times A_2 |) \; bir] \\ c_i.R_i^i + c_j.R_j^i & | \; [ \; (| R_i^i | \bullet | R_j^i |) \; bir] \\ c_i.R_i^o + c_j.R_j^o & | \; [ \; (| R_i^o | \bullet | R_j^o |) \; bir] \end{cases}$$

Theorem 5 and Definition 14 may be applied to a wide range of phenomena in knowledge and intelligence sciences, for instance, the derivation of Theorems 6 and 7. Based on Theorem 5, the general theory of knowledge for both the itemized and entire knowledge may be rigorously described.

**Definition 15.** An *itemized knowledge* $\kappa^0$ in the universe of discourse of knowledge $\mathcal{U}$ is an $(n+1)$-*ary* relation $r_k$ between a particular concept $C_0$ and a set of existing concepts $\underset{i=1}{\overset{n}{R}} C_i$ in a knowledge base:

$$\kappa \triangleq r_k : C_0 \to \underset{i=1}{\overset{n}{R}} C_i, \; C_i \in \mathcal{K} \tag{21}$$

$$= C_0 \underset{i=1}{\overset{n}{X}} C_i = \underset{i=1}{\overset{n+1}{X}} C_i \; [bir]$$

The mathematical model of itemized knowledge may be extended to that of the entire knowledge as a Cartesian product of all formal concepts in the knowledge base of a person or an *SAS*.

**Theorem 6.** The *entire knowledge* $\mathfrak{K}$ is determined by a recursive Cartesian product among all formal concepts $C_i \sqsubseteq \mathfrak{K}$ at all $k$th layers represented by a hierarchical knowledge base:

$$\mathfrak{K} \triangleq \underset{k=1}{\overset{p}{R}} \mathfrak{K}^k (\mathfrak{K}^{k-1}), \quad \mathfrak{K}^0 = \kappa = \underset{i=1}{\overset{n}{X}} C_i \qquad (22)$$

$$= \mathfrak{K}^p (\mathfrak{K}^{p-1} (...(\mathfrak{K}^1 (\underset{i=1}{\overset{n}{X}} C_i))...)) \ [bir]$$

*Proof:* Theorem 6 is inductively proven by Definitions 14 and 15 based on Theorem 1:

$$\mathfrak{K} = \begin{cases} \mathfrak{K}^1 = \kappa = (\underset{i=1}{\overset{n}{X}} C_i) \\ \mathfrak{K}^2 = \mathfrak{K}^1(\mathfrak{K}^0) = \mathfrak{K}^1(\underset{i=1}{\overset{n}{X}} C_i) \\ ... \\ \mathfrak{K}^p (\mathfrak{K}^{p-1}(...(\mathfrak{K}^1(\underset{i=1}{\overset{n}{X}} C_i))...)) \end{cases} \qquad (23)$$

$$= \mathfrak{K}^p (\mathfrak{K}^{p-1}(...(\mathfrak{K}^1(\underset{i=1}{\overset{n}{X}} C_i))...)) = \underset{k=1}{\overset{p}{R}} \mathfrak{K}^k (\mathfrak{K}^{k-1}) \ [bir]$$

∎

The formal model of human knowledge leads to another discovery for explaining how many *bir*s of knowledge that an individual may remember and what the maximum potential of human memory capacity is according to the SAS theory.

**Theorem 7.** The *capacity of human memory* $\Omega(\mathfrak{K})$ for knowledge representation and memorization is bounded by $10^{8432}$ [*bir*] as a result of all potential synaptic connections $n_\psi$ created by the total number of nerves $n_\mu$ in the brain:

$$\Omega(\mathfrak{K}) \leq 10^{8,432} \ [bir], \ \exists n_\mu = 10^{11} \text{ and } n_\psi = 10^3 \qquad (24)$$

*Proof.* Let the total number of neurons and the average synaptic connections among them be $n_\mu = 10^{11}$ and $n_\psi = 10^3$, respectively, based on observations in neurology [21]. $\Omega(\mathfrak{K})$ is determined by all potential synaptic connections in the brain:

$$\exists n_\mu = 10^{11}, \ n_\psi = 10^3,$$

$$\Omega(\mathfrak{K}) \triangleq C_{n_\mu}^{n_\psi} = \frac{n_\mu !}{n_\psi ! \ (n_\mu - n_\psi)!} \qquad (25)$$

$$= \frac{10^{11}!}{10^3 ! \ (10^{11} - 10^3)!}$$

$$\not> 10^{8,432} \ [bir]$$

∎

Theorem 7 indicates that human memory capacity as an *SAS* based on the symbiotic neural structures of the brain is much greater than any human-made memory chip and hard disk, even their collected sum. This reveals that the human memory in the brain may never be exhausted.

The social implication of machine knowledge learning as a brain-inspired *SAS* is represented by an unpresented advantage of symbiotic learning sharable by humans and machines where acquired knowledge may be directly transferred to peers based on the common platform of knowledge representation and the shared knowledge manipulation engines. In this novel approach, the human learning redundancy where similar knowledge needs to be repetitively learned and relearned by individuals may be avoided as enabled by symbiotic learning. Therefore, *SAS*-based machine knowledge learning will become an indispensable form of learning between human and machines by mutually sharable knowledge bases in an unprecedented hybrid *SAS* society. The sixth form of *SAS* learning for knowledge acquisition will enable collective knowledge acquisition for exponentially extending human learning ability toward a wide range of novel applications.

The theoretical framework of symbiotic knowledge learning may trigger the *seventh form* of AI learning discovered recently as introspective learning. *Introspective learning* (*IL*) [34] recursively generates new knowledge based on existing knowledge in the brain, which incubates deepened understanding across previously acquired knowledge in recursive and inductive ways. *IL* happens in the brains and everyday lives as the most advanced and highly frequent form of knowledge learning. The mechanism of *IL* is formally explained in Definition 15 and Theorem 6, which reveals that most of our advanced knowledge is created by *IL* in human brains. According to Theorem 6, *IL* may be implemented by *SAS* technologies supported by a *Cognitive Knowledge Base* (*CKB*) [34] in computational intelligence.

The formal and comprehensive approach described in this paper has revealed the universe of discourse of *SAS* theories and engineering applications, which may lead to groundbreaking developments in the near future. Towards *SAS* implementations and applications, a proper way may be to tailor the theoretical models for specific problems along the line of the human-in-the-loop paradigm. In addition to the example of *SAS* for symbiotic knowledge learning, many *SAS* paradigms are emerging encompassing: 1) Brain-inspired systems [12]; 2) Cognitive robots [31]; 3) Intelligent IoT ($I^2oT$) [32]; 4) Social computing systems; 5) Brain-machine interfaces; 6) Cognitive systems; 7) Bioinformatics systems; 8) Internet of Minds (IoM); 9) Self-driving automobiles; 10) Human-collaborative robots; 11) Unmanned systems; 12) Manned-unmanned teaming; 13) Robot memetics [47]; and 14) Intelligent defence systems. It is noteworthy that none of the *SAS* applications is trivial towards the next generation of cognitive computers, *GAI*, and hybrid symbiotic human-machine societies. Related *SAS* and *AS* projects undertaken in our labs address challenges for abstract intelligence, intelligent mathematics for *SAS*, the tripartite framework of SAS trustworthiness, autonomous decision making, a transdisciplinary theory for cognitive cybernetics, humanity, and systems science, cognitive foundations of knowledge science, and the abstract system theory for *SAS*. The advances of *SAS* theories and technologies should lead from information revolution to the era of intelligence revolution for unprecedented breakthroughs for enabling pervasive human-machine symbiotic systems.

## 5. Conclusion

This work has developed a theoretical framework of symbiotic autonomous systems (*SAS*) as an emerging field of advanced and general AI technologies. The architecture of *SAS* has been reduced to the synergy between symbiotic and autonomous systems. Extended intelligent behaviors of *SAS* have been formally revealed by the hierarchical intelligence model (HIM). The fundamental challenge to seamlessly enable human and machine interactions in a hybrid environment has been formally addressed. This work has revealed that: a) *SAS* are a recursive structure of intelligent systems; b) *SAS* are collective intelligent systems for human societies and engineering applications; c) *SAS* represent a general AI technology that leads to emerging hybrid societies of human and intelligent machines; and d) *SAS* are underpinned by the advances in intelligent mathematics for enabling advanced epistemology learning by machines. The basic research and empirical technologies presented in this work have provided a rigorous foundation towards *SAS* modeling and implementations, which may trigger a wide range of novel applications for augmenting human capabilities and intelligent power. *SAS* are expected to open a new era from information revolution to intelligence revolution.

## Acknowledgment

This work is supported in part by the Canadian Department of National Defence through the AutoDefence project, Natural Sciences and Engineering Research Council (NSERC), and the IEEE SMC Society Technical Committee on Brain-Inspired Cognitive Systems (TC-BCS). The authors would like to thank the anonymous reviewers for their valuable suggestions and comments.

## References

[1] I. Newton (1729), *The Principia: The Mathematical Principles of Natural Philosophy*, Benjamin Motte, London.
[2] T. O'connor and D. Robb Eds., *Philosophy of Mind: Contemporary Readings*, Routledge, London, UK, 2013.
[3] S. Coradeschi and A. Saffiotti (2006), Symbiotic Robotic Systems: Humans, Robots, and Smart Environments, *IEEE Intelligent Systems*, 21(3), pp. 82–84.
[4] M.E. Crosby, J. Scholtz, and P. Ward (2006), Symbiotic Performance between Humans and Intelligent Systems, *Journal of Interacting with Computers*, 18(6), pp. 1165–1169.
[5] Y. Wang (2018), Keynote: The Emergence of Abstract Sciences and Brain-inspired Symbiotic Systems, *IEEE Future Direction Committee Workshop on Symbiotic Autonomous Systems* (WSAS'18), Miyazaki, Japan, Oct., p. 2.
[6] S. Doltsinis, P. Ferreira, and N. Lohse (2018), A Symbiotic Human–Machine Learning Approach for Production Ramp-up, *IEEE Transactions on Human-Machine Systems*, 48(3), pp. 229–240.

[7] Y. Wang and E. Tunstel (2019), Emergence of Abstract Sciences and Transdisciplinary Advances in Systems, Man, and Cybernetics, *IEEE System, Man and Cybernetics Magazine*, 5(2), pp.12–19.

[8] Y. Wang, M. Hou, K.N. Plataniotis, S. Kwong, H. Leung, E. Tunstel, I.J. Rudas, and L. Trajkovic (2021), Towards a Theoretical Framework of Autonomous Systems Underpinned by Intelligence and Systems Sciences, *IEEE/CAS Journal of Automatica Sinica*, 8(1):52–63.

[9] V. Mnih et al.(2015), Human-level Control through Deep Reinforcement Learning, *Nature*, 518:529–533.

[10] M. Hou, S. Banbury, and C. Burns (2014), *Intelligent Adaptive Systems: An Interaction-Centered Design Perspective*, CRC Press, Boca Raton, FL., USA.

[11] D.P. Watson and D.H. Scheidt (2005), Autonomous Systems, *Johns Hopkings Appl. Tech. Digest*, 26(4), pp. 268–376.

[12] Y. Wang, S. Kwong, H. Leung, J. Lu, M.H. Smith, L. Trajkovic, E. Tunstel, K.N. Plataniotis, G. Yen, and W. Kinsner (2020), Brain-Inspired Systems: A Transdisciplinary Exploration on Cognitive Cybernetics, Humanity, and Systems Science towards AI, *IEEE System, Man and Cybernetics Magazine*, 6(1):6–13.

[13] Y. Wang, S. Yanushkevich, M. Hou, K.N. Plataniotis, M. Coates, M. Gavrilova, Y. Hu, F. Karray, H. Leung, A. Mohammadi, S. Kwong, E. Tunstel, L. Trajkovic, I.J. Rudas, and J. Kacprzy (2020), A Tripartite Framework of Trustworthiness of Autonomous Systems, *IEEE 2020 Int'l Conf. Systems, Man, and Cybernetics*, Oct. pp. 3375-3380.

[14] H.A. Abbass, G. Leu, and K. Merrick (2016), A Review of Theoretical and Practical Challenges of Trusted Autonomy in Big Data, *IEEE Access*, vol. 4, pp. 2808–2830.

[15] Y. Wang (2020), A Rigorous Cognitive Theory for Autonomous Decision Making, *IEEE 2020 Int'l Conf. on Systems, Man, and Cybernetics* (SMC'20), Toronto, Oct. pp. 1021–1026.

[16] E.A. Bender (2000), *Mathematical Methods in Artificial Intelligence*, IEEE CS Press, Los Alamitos, CA.

[17] G.J. Klir (1992), *Facets of Systems Science*, Plenum, NY.

[18] Y. Wang (2015), A Denotational Mathematical Theory of System Science: System Algebra for Formal System Modeling and Manipulations, *Journal of Advanced Mathematics & Applications*, 4(2):132–157.

[19] M. Bunge (1978), General Systems Theory Challenge to Classical Philosophy of Science, *Int. J. Gen. Systems,* 4(1), 3–28.

[20] Y. Wang (2014), Software Science: On General Mathematical Models and Formal Properties of Software, *Journal of Advanced Mathematics & Applications*, 3(2), 130–147.

[21] R.A. Wilson and C.K. Frank Eds., *The MIT Encyclopedia of the Cognitive Sciences*, MIT Press, MA, 2001.

[22] Y. Wang (2020), Keynote: Intelligent Mathematics (IM): Indispensable Mathematical Means for General AI, Autonomous Systems, Deep Knowledge Learning, Cognitive Robots, and Intelligence Science, *IEEE 19th Int'l Conf. on Cognitive Informatics & Cognitive Computing* (ICCI*CC'20), Tsinghua Univ., Beijing, China, Sept., p.5.

[23] Y. Wang (2012) , In Search of Denotational Mathematics: Novel Mathematical Means for Contemporary Intelligence, Brain, and Knowledge Sciences, *Journal of Advanced Mathematics and Applications*, 1(1):4–25.

[24] Y. Wang, H. Leung, M.L. Gavrilova, O. Zatarain, D. Graves, J. Lu, N. Howard, S. Kwong, P. Sheu, and S. Patel (2018), A Survey and Formal Analyses on Sequence Learning Methodologies and Deep Neural Networks, *17th IEEE Int'l Conf. Cognitive Informatics & Cognitive Computing* (ICCI*CC'18), Univ. of California, Berkeley, USA, July, pp. 6–15

[25] C.L.P. Chen, Z. Liu, and S. Feng (2020), Universal Approximation Capability of Broad Learning System and Its Structural Variations, *IEEE Trans. On Neural Networks and Learning Systems*, April, 30(4):1191–1204.

[26] Y. Wang (2018), Keynote: Cognitive Foundations and Formal Theories of Human and Robot Visions, *17th IEEE Int'l Conf. Cognitive Informatics & Cognitive Computing* (ICCI*CC 2018), UC Berkeley, USA, July, p. 5.

[27] J.W. Lu, K.N. Plataniotis, and A.N. Venetsanopoulos (2003), Face Recognition using Kernel Direct Discriminant Analysis Algorithm, *IEEE Trans. on Neural Networks*, 14(1):117–126.

[28] Y. Wang (2016), "Deep learning and deep reasoning by cognitive robots and computational intelligent systems," *Proc. 15th IEEE Int. Conf. Cognitive Informatics & Cognitive Computing* (ICCI*CC'16), Stanford Univ., July, p. 3.

[29] Y. Wang (2017), Keynote: Cognitive Foundations of Knowledge Science and Deep Knowledge Learning by Cognitive Robots, *16th IEEE Int'l Conf. Cognitive Informatics & Cognitive Computing* (ICCI*CC 2017), Univ. of Oxford, UK, IEEE CS Press, July, pp. 4.

[30] C.E. Shannon (1948), A Mathematical Theory of Communication, *Bell System Tech. J.,* 27, 379-423, 623-656.

[31] Y. Wang (2010), Cognitive Robots: A Reference Model towards Intelligent Authentication, *IEEE Robotics and Automation*, 17(4):54–62.

[32] G. Fortino, W. Russo, C. Savaglio, W. Shen and M. Zhou (2018), Agent-Oriented Cooperative Smart Objects: From IoT System Design to Implementation, *IEEE Transactions on Systems, Man, and Cybernetics: Systems*, Nov., 48(11):1939–1956.

[33] Y. Ma, Z. Wang, H. Yang and L. Yang (2020), "Artificial intelligence applications in the development of autonomous vehicles: a survey," *IEEE/CAA Journal of Automatica Sinica*, March, 7(2):315–329.

[34] Y. Wang (2014), On a Novel Cognitive Knowledge Base (CKB) for Cognitive Robots and Machine Learning, *Int'l J. Software Scie. & Comp. Intell.*, 6(2):42-64.

[35] M. Song, D. Tao, Z. Liu, X. Li and M. Zhou (2010), "Image Ratio Features for Facial Expression Recognition Application," *IEEE Transactions on Systems, Man, and Cybernetics*, Part B (Cybernetics), June, 40(3):779–788.

[36] S. Yang, Y. Wen, L. He and M. Zhou (2020), "Sparse Common Feature Representation for Undersampled Face Recognition," *IEEE Internet of Things Journal*, doi: 10.1109/JIOT.2020.3031390.

[37] J. Kacprzyk, R.R. Yager and J.M. Merigo (2019) Towards Human-Centric Aggregation via Ordered Weighted Aggregation Operators and Linguistic Data Summaries: A New Perspective on Zadeh's Inspirations, *IEEE Computational Intelligence Magazine,* 14(1): 16–30.

[38] J. Kacprzyk, H. Nurmi, and S. Zadrozny (2017), Reason vs. Rationality: From Rankings to Tournaments in Individual Choice, *Trans. Computational Collective Intelligence*, vol. 27, pp. 28–39.

[39] A.W. Woolley, C.F. Chabris, A. Pentland, N. Hashmi, and T.W. Malone (2010), Evidence of a Collective Intelligence Factor in the Performance of Human Groups, *Science*, 330(6004):686–688.

[40] Y. Wang (2020), Keynote: *How will Autonomous Systems and Cognitive Robots Augment Human Intelligence? Future Technologies Conference* (FTC'20), Vancouver, Canada, Nov. p. 2.

[41] M. Martin, C. Lebiere, M. Fields et al. (2020), Learning features while learning to classify: a cognitive model for autonomous systems. *Comp. Math Organ Theory*, 26, pp. 23–54.

[42] T.J.M. Bench-Capon (2020), Ethical approaches and autonomous systems, *Artificial Intelligence*, 281, pp.103239.

[43] E. Netzer, A.B. Geva (2020), Human-in-the-loop active learning via brain computer interface. *Ann Math Artif. Intell.,* 88, pp.1191–1205.

[44] G. Miriam, M. Albert, J. Fons, and V. Pelechano (2020), Engineering human-in-the-loop interactions in cyber-physical systems, *Information and Software Technology*, 126, pp.106349.

[45] Y. Wang (2008), On Cognitive Properties of Human Factors and Error Models in Engineering and Socialization, *Int'l J. of Cogn. Infor. and Nat'l Intell.,* 2(4), Oct., 70-84.

[46] E. Tunstel, M.J. Cobo, E. Herrera-Viedma, I.J. Rudas, D. Filev, L. Trajkovic, C.L.P. Chen, W. Pedrycz, M.H. Smith, and R. Kozma (2021), Systems Science and Engineering Research in the Context of Systems, Man, and Cybernetics: Recollection, Trends, and Future Directions, *IEEE Trans. Systems, Man, and Cybernetics: Systems* (to be published).

[47] W. Truszkowski, C. Rouff, M. Akhavannik, and E. Tunstel (2020), *Robot Memetics: A Space Exploration Perspective*, Springer Briefs in Electrical and Computer Engineering, Springer International Publishing.

[48] R. Saracco (2017), *Looking Ahead to 2050 - Symbiotic Autonomous Systems ( I-XI)*, IEEE Future Directions Blog, https://digitalreality.ieee.org/publications/blog/looking-ahead-to-2050.